\documentclass{article}

\usepackage[preprint,nonatbib]{neurips_2023}

\usepackage[utf8]{inputenc} 
\usepackage[T1]{fontenc}    
\usepackage{hyperref}       
\usepackage{url}            
\usepackage{booktabs}       
\usepackage{amsfonts}       
\usepackage{nicefrac}       
\usepackage{microtype}      
\usepackage{xcolor}         

\usepackage[sorting=none,maxbibnames=99]{biblatex}
\usepackage{amsmath}
\usepackage{graphicx}
\usepackage{subcaption}
\usepackage{adjustbox}
\usepackage{mathtools}
\usepackage{caption} 
\title{Siamese Networks with Soft Labels for Unsupervised Lesion Detection and Patch Pretraining on Screening Mammograms}

\begin{document}

\author{%
  Kevin Van Vorst \\
  Icahn School of Medicine at Mount Sinai\\
  New York, NY 10029-5674 \\
  \texttt{kevin.vanvorst@icahn.mssm.edu}
  \And
  Li Shen \\
  Icahn School of Medicine at Mount Sinai \\
  New York, NY 10029-5674 \\
  \texttt{li.shen@mssm.edu}
  }
\maketitle

\begin{abstract}
Self-supervised learning has become a popular way to pretrain a deep learning model and then transfer it to perform downstream tasks. However, most of these methods are developed on large-scale image datasets that contain natural objects with clear textures, outlines, and distinct color contrasts. It remains uncertain whether these methods are equally effective for medical imaging, where the regions of interest often blend subtly and indistinctly with the surrounding tissues. In this study, we propose an alternative method that uses contralateral mammograms to train a neural network to encode similar embeddings when a pair contains both normal images and different embeddings when a pair contains normal and abnormal images. Our approach leverages the natural symmetry of human body as weak labels to learn to distinguish abnormal lesions from background tissues in a fully unsupervised manner. Our findings suggest that it's feasible by incorporating soft labels derived from the Euclidean distances between the embeddings of the image pairs into the Siamese network loss. Our method demonstrates superior performance in mammogram patch classification compared to existing self-supervised learning methods. This approach not only leverages a vast amount of image data effectively but also minimizes reliance on costly labels, a significant advantage particularly in the field of medical imaging.
\end{abstract}

\section{Introduction}

The creation of large image databases such as the ImageNet \cite{5206848} has made it possible to develop powerful artificial neural networks (ANNs) with millions of parameters to classify images at very high accuracy. This has revolutionized computer vision where the use of large-scale ANNs, known as deep learning, has become standard practice \cite{dl}. It has also resumed people's interest in developing the next-generation computer-aided diagnosis (CAD) tools in medical imaging \cite{cad}, where the progress has stagnated for decades since 1990s. However, unlike natural image datasets that can be labeled through crowd-sourcing \cite{crowdsource}, medical image datasets are notoriously expensive and time consuming to create. They require qualified experts, whose times are often constrained, to verify these images are correctly labeled \cite{preparemedimg}. To make the problem even worse, there is often a significant amount of variability among the experts \cite{doi:10.1148/ryai.220056}. 

A major theme in machine learning is to teach models to learn from unlabeled data through unsupervised learning. In recent years, a family of unsupervised learning methods known as self-supervised learning (SSL) has emerged as a highly effective way of learning without labels. In a nutshell, SSL generates artificial tasks from data for a model to solve, through which the model learns to extract meaningful representations from the data \cite{ssl_review}. This process is known as pretraining. A pretrained model becomes an encoder whose outputs can be directly used or finetuned for downstream tasks, often with much less supervision than a model that is learned from scratch \cite{ssl_general}. SSL has proved to be successful in medical imaging tasks \cite{ssl_med}. We have previously used SSL on mammographic images to train a model that reaches an accuracy nearly as high as a fully supervised model using only 25\% of the labels in breast cancer detection \cite{miller}.

A distinctive feature of medical images is that they are often taken from human body parts that are naturally symmetrical. This symmetry can potentially serve as a form of weak labels that can be leveraged to teach models to learn features that can classify abnormal and normal samples from inputs. In this study, we propose an alternative to the SSL methods to exploit the symmetry for representation learning without explicit labels. Our models are trained on bilateral mammogram patch pairs to encode similar embeddings when both patches of a pair are normal and different embeddings when one of the patches is abnormal without being given the labels of the patches. We show that this objective can be formulated as the loss of a Siamese network with soft labels. We then show the effectiveness of our models on several downstream tasks in comparison to the SSL methods.

\section{Related Works}
\subsection{Self Supervised Learning Methods}

SSL is a class of machine learning methods where a model is trained on unlabeled data to learn general and useful representations \cite{ssl_general}. The pretrained model can then be used as an encoder to extract embeddings for downstream tasks. Generally speaking, SSL methods can be classified into two groups: pretext tasks and contrastive learning \cite{sslreview}. Learning representations via pretext tasks involves generating pseudo labels, e.g. via rotation, masking, or colorization, and ask the model to predict the generated labels \cite{sslreview}. On the other hand, contrastive learning does not use pseudo labels, but rather applies strong data augmentation to a single image $p$ to produce two distorted views using a stochastic transformation function $t$ so that $v_1=t(p), v_2=t(p)$. The two views from the same image are called a positive pair while two views from two different images are called a negative pair. In SimCLR \cite{simclr}, an encoder is trained to maximize the agreement of positive pairs and simultaneously minimize the agreement of negative pairs. Another popular SSL method is called BYOL \cite{byol} where only positive pairs are used. In BYOL, an online network $f$ is learned to encode views and a target network $g$ is created as an exponential moving average of the online network. The learning task is to maximize the agreement between the online and target networks' representations $f(v_1)$ and $g(v_2)$. In our previous work \cite{miller}, we found both methods to be effective in learning representations from mammographic images for breast cancer detection. In this work, our focus is on a bilateral patch pair $(p_1, p_2)$ that comes from the two breasts of the same patient. However, the learning objective is in spirit somewhat similar to contrastive learning in the sense that we want to maximize the agreement when $(p_1, p_2)$ is a normal pair (i.e., both $p_1$ and $p_2$ are normal) and minimize the agreement when $(p_1, p_2)$ is an abnormal pair (i.e., either $p_1$ or $p_2$ is abnormal).

\subsection{Siamese Networks}

Siamese networks are a class of neural network architectures that consist of two identical networks with shared weights \cite{origsiam} but they work on two different inputs to compute comparable outputs. For a pair of input images $(p_1, p_2)$, the learning objective is to compute similar representations when $p_1$ and $p_2$ come from the same class and dissimilar representations when they come from different classes. Assume the image encoding part of a Siamese network is represented by function $g$, the embeddings of the pair of images are $h_1=g(p_1)$ and $h_2=g(p_2)$. The Siamese network learning can be setup as a binary classification task on the concatenated embedding $h=concat(h_1, h_2)$ so that $f(h)=q$ represents the probability that the image pair comes from the same class, where $f$ is a binary classifier implemented as a fully connected layer. The binary cross entropy loss can be used to train the Siamese network:
\begin{equation} \label{eq1}
\begin{split}
L = -[y\cdot log(q) + (1-y)\cdot log(1-q)]
\end{split}
\end{equation}
where $y\in\{0,1\}$ is the ground-truth label for the pair to be from the same class. 

Siamese networks were originally developed for facial recognition \cite{facenet} and later found success in other areas such as cancer prediction in chronologically paired mammogram images \cite{featurefusion}. In this study, we use a Siamese network to encode a pair of patches $(p_1, p_2)$ from bilateral mammograms. If the pair is normal we treat it as from the same class; if it is abnormal we treat it as from different classes. However, \textit{we would not know if a pair is normal or abnormal for an unlabeled dataset}. We deal with that by introducing \textit{soft labels} into the loss function.

\subsection{Label Noise Modeling}

Label noise learning refers to training models on data that contain corrupted labels \cite{han2020survey}. This problem reflects real world scenarios where samples are mislabeled or missing labels. Many techniques have been developed to deal with label noise. One class of methods uses mixture modeling to identify mislabeled samples \cite{ICML2019_UnsupervisedLabelNoise,dividemix} based on two premises: 1. Despite noisy labels, a model can still learn to somewhat classify samples correctly based on the clean samples; 2. Mislabeled samples tend to have greater losses than clean samples. Consequently, the samples can be separated into ``noisy" and ``clean" groups based on losses as soon as the model is reasonably trained.

Inspired by the mixture modeling method in label noise learning, we use Gaussian mixture models (GMMs) to identify abnormal pairs from normal pairs in an unsupervised manner. Although the patch pairs have unknown labels to begin with, as the Siamese network learns to compute representations for a patch pair, an abnormal pair tends to contain representations that are less similar than a normal pair. This provides an opportunity to distinguish them using an unsupervised clustering technique.

\section{Methods}
\subsection{Soft Label and Gaussian Mixture Modeling}

Our aim is to identify abnormal patch pairs from bilateral mammograms in an unsupervised manner. Since the true label of any given patch pair generated from a pair of bilateral mammograms is unknown, we introduce a “soft” label, $P\in[0,1]$, to represent the confidence for the patch pair being abnormal. Assuming a neural network model has already been learned to encode patches in a sensible way for an abnormal patch to distinguish from its paired normal patch, then the Euclidean distance between the embeddings of an abnormal pair should be higher than that of a normal pair. Let function $g$ represent the part of the network up to the embedding layer, the embeddings for the patch pair $(p_1,p_2)$ are $e_1=g(p_1),e_2=g(p_2)$. The Euclidean distance is defined as $D=d(e_1,e_2)$. A GMM $h$ can be built on the set of Euclidean distances for all patch pairs on the training set defined as $C=\{D_i\}, i=1..N$. Here, a two-component GMM is fit on $C$ to identify the two classes (abnormal vs. normal) of patch pairs. The GMM function $h$ can be used to provide the posterior probability that a pair with distance $D$ belongs to the abnormal class such that the soft label $P=h(D)$.

We used the Python package sklearn \cite{scikit-learn} to handle GMM fitting and posterior probability scoring. After fitting the GMM, it can be used to predict $P$ for each patch pair, which will be used as the soft labels in the loss function to be introduced below.

\subsection{Proposed Model}
The proposed model is shown in Figure 1 where a Siamese network is used to classify a patch pair. ResNet-18 \cite{resnet18} is used as the image encoder with the global average pooling layer used as the embedding. The two embeddings $(e_1,e_2)$ from a patch pair $(p_1,p_2)$ is concatenated so that $E=concat(e_1,e_2)$. The concatenated embedding is passed through a fully connected layer with sigmoid activation $f$ to a single output node to predict the class of the patch pair. This results in $q=f(E)\in[0,1]$ representing the probability that the patch pair is normal.

As described in the previous section, the embedding pair $(e_1,e_2)$ is also used to derive soft label $P$ based on Euclidean distances $D=d(e_1,e_2)$ across the training set. However, there is a significant distinction between soft label $P$ and probability $q$. $q$ represents the probability that $(p_1,p_2)$ is a normal pair directly computed by the Siamese network, while $P$ represents the ``label" that $q$ tries to match and is derived from an unsupervised clustering method trained on the entire training set. By replacing the ground truth label $y$ with soft label $P$ ($1-P$ for $y$) in the binary cross entropy loss (eq.1), we have: 
\begin{equation} \label{eq2}
\begin{split}
L = -[(1-P)\cdot log(q) + P\cdot log(1-q)]
\end{split}
\end{equation}

\begin{figure}
    \centering
    \includegraphics[scale=0.25]{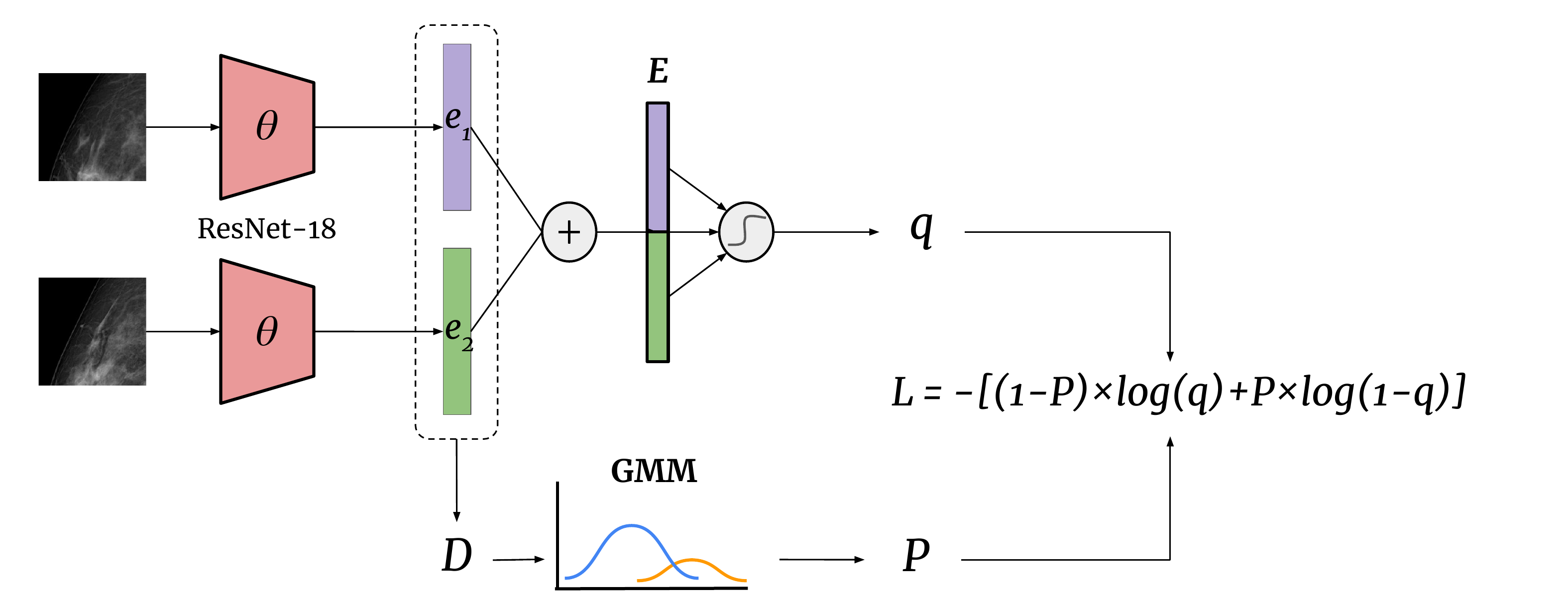}
    \caption{Two parallel networks, with shared weights \(\theta\), process a pair of patches and return embeddings $e_1$ and $e_2$. The Euclidean distance, $D=d(e_1,e_2)$, is calculated. A two-component GMM is fit on the $D$ from the entire training set to get $P$. The two embeddings are concatenated to single vector $e$ and passed through a fully connected layer with sigmoid activation to get $q$.}
    \label{fig:singlesiamese}
\end{figure}

Initial experiments showed that training such a network was unstable. Multiple runs using the same parameters can result in different performances. This might be the result of confirmation bias where an initial wrong guess can be amplified and confirmed repeatedly throughout training. To deal with this instability, a second Siamese network is trained simultaneously where the $q$ and $P$ from the two networks are cross used in each other's losses. This idea is inspired by the work of DivideMix \cite{dividemix}. Let the soft label and normal pair probability from Siamese network 1 be $P_1, q_1$ and Siamese network 2 be $P_2, q_2$, the losses for the two networks are now:
\begin{equation} \label{eq3}
\begin{split}
L_1 = -[(1-P_2)\cdot log(q_1) + P_2\cdot log(1-q_1)]
\end{split}
\end{equation}
\begin{equation} \label{eq4}
\begin{split}
L_2 = -[(1-P_1)\cdot log(q_2) + P_1\cdot log(1-q_2)]
\end{split}
\end{equation}
\noindent
Figure 2 shows the interaction of the two Siamese networks and GMMs in their respective loss functions. The overall loss is simply the average of $L_1$ and $L_2$. We found using two networks greatly improved the learning stability.

\begin{figure}[h]
    \centering
    \includegraphics[scale=0.2]{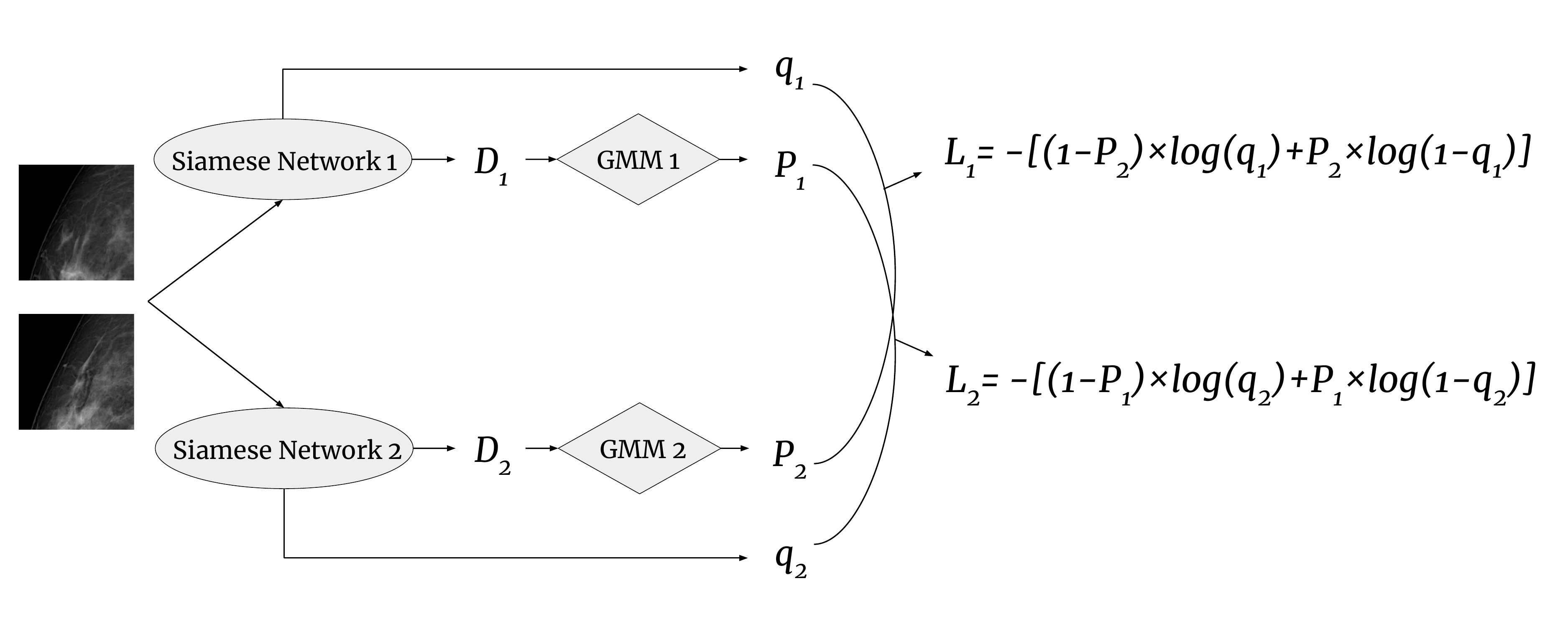}
    \caption{A pair of mammogram patches is encoded by two separate Siamese networks resulting in normal pair probabilities $q_1$ and $q_2$ and Euclidean distances $D_1$ and $D_2$. The two Euclidean distances are used in their respective GMMs to get the soft labels $P_1$ and $P_2$.}
    \label{fig:doublesiamese}
\end{figure}

\section{Datasets}
Two datasets are used in this study: VinDr-Mammo \cite{Nguyen2022.03.07.22272009} and OPTIMAM \cite{OPTIMAM}. The VinDr-Mammo dataset contains 20,000 images from 5,000 mammography studies with radiologists' assessment and lesion annotations but without further confirmation of cancer status. Each study contains the bilateral images of both the craniocaudal (CC) and mediolateral oblique (MLO) views. Every exam was double read receiving a BI-RADS assessment and bounding box coordinates with any disagreement settled by the opinion of a third radiologist. 6,703 exams was assigned to category 1 (negative), 2,338 to category 2 (benign), 465 to category 3 (probably benign), 381 to category 4 (suspicious), and 113 to category 5 (highly suggestive of malignancy). Exams that received a rating of categories 2-5 and were missing bounding box coordinates were dropped. 
\newline

The OPTIMAM dataset is a large-scale database from the United Kingdom with studies from 172,282 patients of mammography images containing annotations and other clinical information including pathology confirmed cancer status. We obtained access to the "standard" subset of this dataset containing 18,898 patients. Due to the massive amount of patient data and potential image patches that can be made, 1,000 patients with screening cases containing images of only CC and MLO views were randomly chosen to be used for pretraining. Each case had a final outcome assignment of either normal (N), malignant finding (M), malignant finding with annotations (M+), benign finding (B), or benign finding with annotations (B+). In this selection, 750 patients have an outcome of N, 104 were M+, 69 were B+, 41 were B, and 36 were M. 

All images in both datasets were resized to 1152x896 and saved as 16-bit unsigned integer PNG files. 

\subsection{Patch Pair Creation}

Since the goal of our approach is to train networks in an unsupervised manner, a uniform grid sampling strategy is used to generate patches from whole mammograms without regard to lesion annotations. Directly training with whole mammograms is computationally prohibitive. Applying random transformation such as cropping on whole mammograms may also accidentally remove lesions, making an abnormal mammogram become normal. Additionally, a lesion’s size is only a fraction of the size of an entire mammogram. Using patches allows models to pay attention to the features of these lesions in greater detail. For these reasons, we only used patches for pretraining. This is the same strategy adopted in a previous study \cite{miller}. 
\par Before sampling patches from a pair of bilateral mammograms, the two images need to be aligned with each other. This can be done through image registration using the Python package SimpleITK \cite{simpleitk}. First, bilateral images of the same view, CC or MLO, are registered to each other by flipping one image and aligning it with the other. For abnormal pairs, the image with no region of interests (ROIs) is always the registered image in order to avoid needing to alter the bounding box coordinates. After image registration, both images are then split into patches in a uniform grid fashion. We sampled square patches of sizes $96\times96$ and $256\times256$. A patch pair is defined as a pair of mammogram patches originating from the same location on the grid. Patch pairs containing more than 50\% background pixels or major border disagreement due to image registration were dropped.

Applying this process to both whole image datasets while using different grid patch sizes, results in four patch pair datasets appropriately named by mother dataset and patch size: VinDr-96, VinDr-256, OPTIMAM-96, and OPTIMAM-256. From the VinDr dataset, 31,785 and 214,942 patch pairs of sizes $256\times256$ and $96\times96$, respectively, were sampled. From the OPTIMAM dataset, 47,444 patch pairs of size $256\times256$ were sampled from 1,000 patients; similarly, 492,394 patch pairs of size $96\times96$ from 1,000 patients.
Figure 3 shows an example of an abnormal and normal patch pair from the VinDr-256 paired patch dataset. In Figure 3a this patch pair originates from a case that received a BI-RADS rating of 5 and it's finding is indicated by the red bounding box. All of the four paired patch datasets were split into training, validation, and test sets based on patients at an 8:1:1 ratio.

\begin{figure}
    \centering
    \begin{subfigure}[t]{0.45\textwidth}
        \centering
        \includegraphics[scale=0.2]{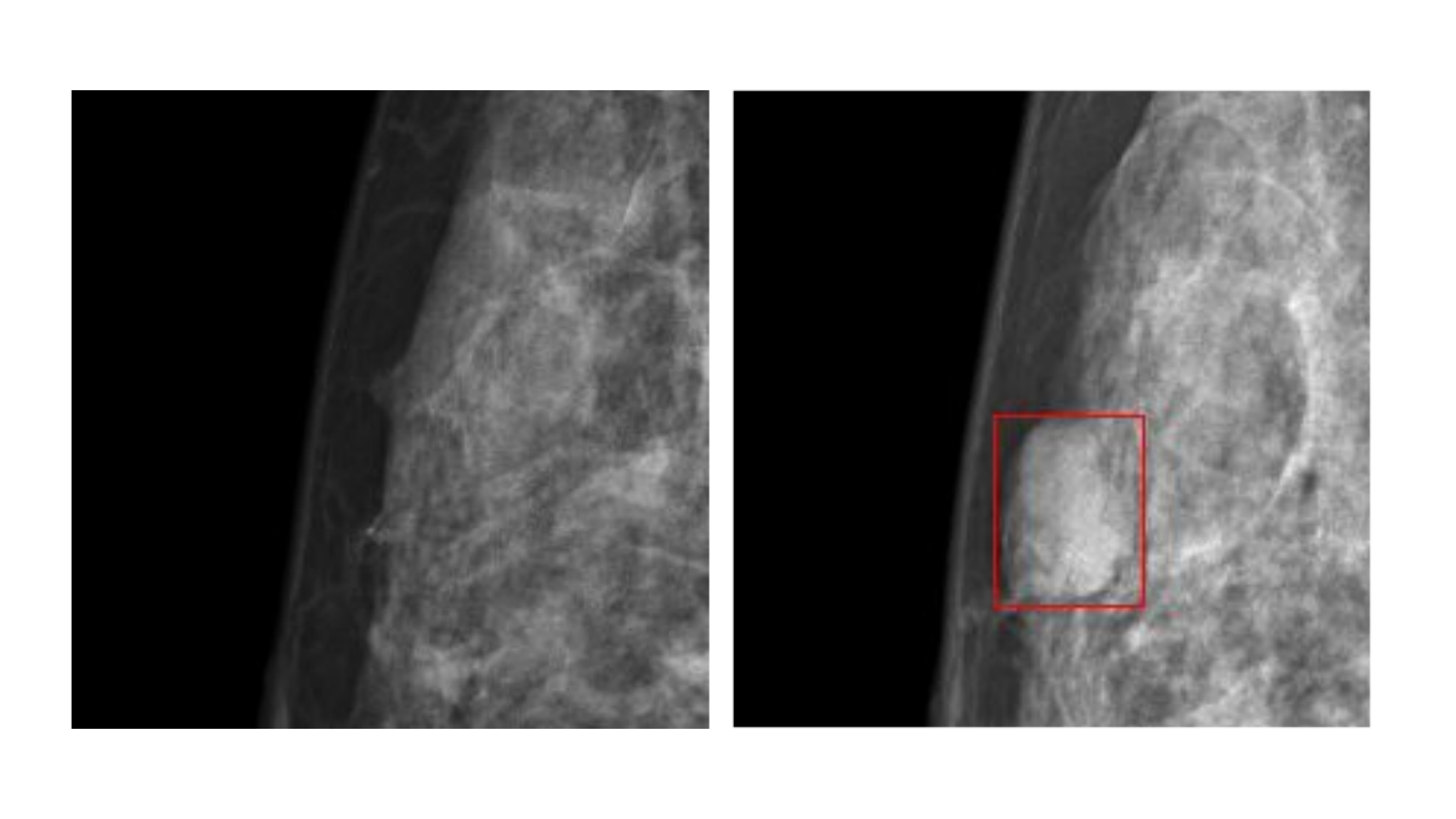}
        \caption{Abnormal patch pair with a mass located in the right patch indicated by a bounding box.}
        \label{fig:abnormalpair}
    \end{subfigure}
    \hfill
    \begin{subfigure}[t]{0.45\textwidth}
        \centering
        \includegraphics[scale=0.2]{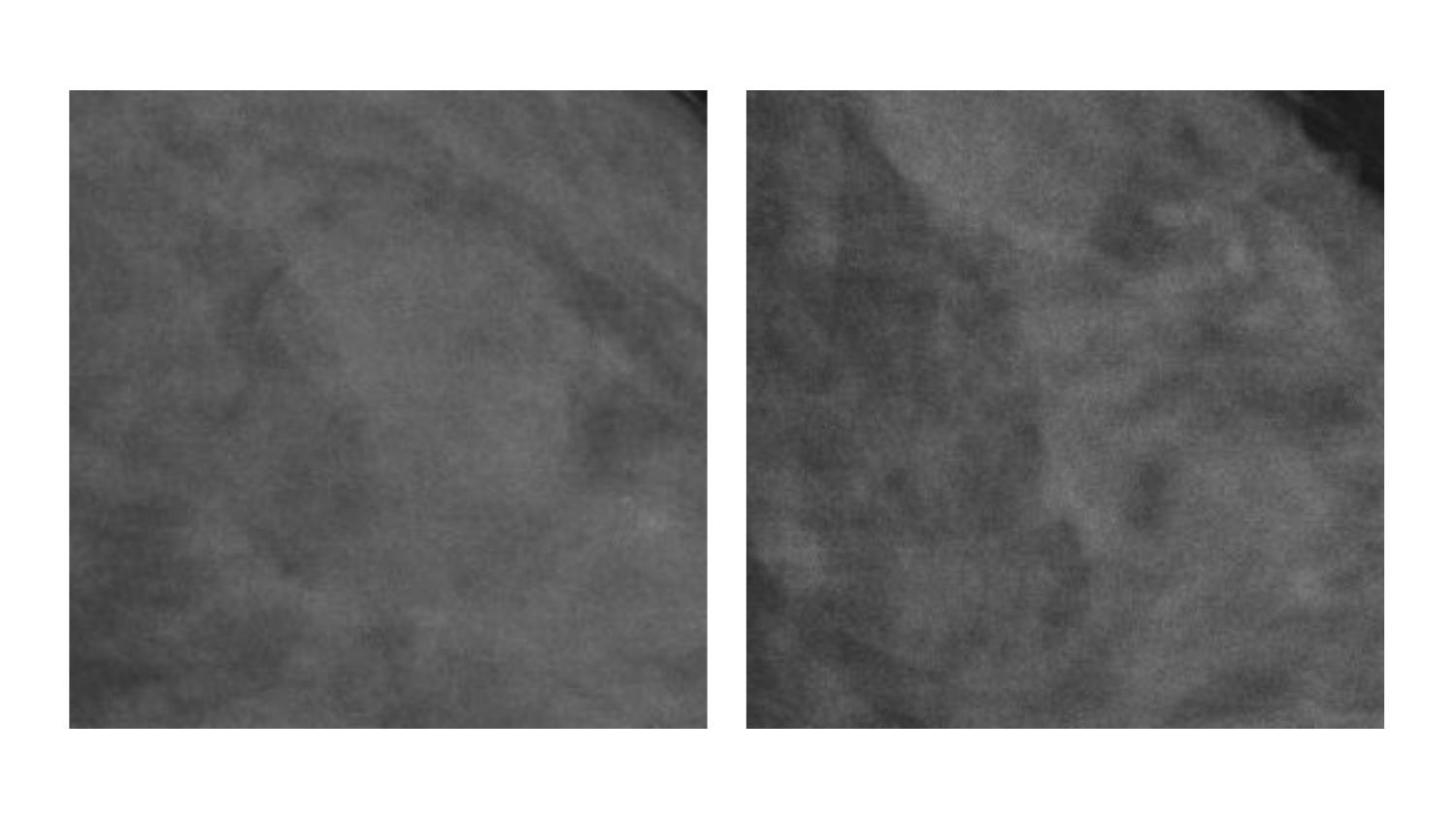}
        \caption{Normal patch pair of background tissues.}
        \label{fig:Normalpair}
    \end{subfigure}
    \label{fig:normalpair}
    \caption{Example patch pairs of size $256\times256$ sampled from the VinDr dataset}
\end{figure}

\subsection{Single Patch Datasets for SSL methods}
Since the SSL methods use only single images as input, the patch pairs created above were split up into individual patches. The SSL patch datasets are 63,570 single patches from the VinDr-256 dataset, 429,884 patches from VinDr-96, 94,888 patches from OPTIMAM-256, and 984,788 patches from OPTIMAM-96. At patch level, all of these datasets are split into training, validation, and test sets at an 8:1:1 ratio.

\subsection{Downstream Task Patch Datasets}

After pretraining, all models are evaluated on several downstream tasks. We created labeled patch datasets for these tasks. This requires sampling the abnormal patches using the bounding box coordinates for each ROI. The abnormal patch of sizes $96\times96$ and $256\times256$ are directly sampled using the center of the ROI. A background patch is sampled from the normal image at the same location as well. These patches are assigned appropriate labels depending on the downstream task. 
The three downstream tasks are the binary classification of abnormal versus normal patches, BI-RADS classification of VinDr patches, and outcome classification of OPTIMAM patches. On the VinDr datasets, the abnormal class is defined as BI-RADS 3-5 and the normal class is defined as BI-RADS 1. We ignored the BI-RADS 2. There are 1,126 patches in both the abnormal and normal classes for a total of 2,252 patches in the labeled VinDr datasets. The breakdown of the BI-RADS labels in the abnormal class are 414 belonging to BI-RADS 3, 453 to BI-RADS 4, and 259 to BI-RADS 5. On the OPTIMAM datasets, the abnormal class is defined as benign and malignant lesions and the normal class is defined as background tissues with no overlap with any ROI. In the available subset, there are 10,981 abnormal patches identified from screening mammograms and 10,981 normal patches containing background tissues. Of the patches in the abnormal class, 1,250 have a benign (B) outcome and 9,731 have a malignant (M) outcome. 

Every dataset is split to training, validation, and test sets at an 8:1:1 ratio at patch level.

\section{Experiments}
\subsection{Siamese Network Patch Pair Training}
Previous SSL methods tend to work better on larger batch sizes \cite{simclr,byol}. We were curious if the Siamese network's performance is also affected by batch size. We performed a grid search on batch sizes, $B \in \{64, 128 ,256, 512, 1028, 2048\}$, and learning rates, \(lr \in \{1.0\times10^{-3}, 1.0\times10^{-4}, 1.0\times10^{-5}, 1.0\times10^{-6}, 1.0\times10^{-7}\)\}, and recorded the validation and test set performances. For every training on this grid search,  the model was trained for 50 epochs at batch size $B$ and two LARS (Layer-wise Adaptive Rate Scaling) \cite{lars} optimizers were used for both Siamese networks with learning rate $lr$. The LARS optimizer uses a separate adaptive learning rate for each layer in the network. We excluded the batch normalization and bias parameters from this layer adaptation. Due to computational resource constraints, gradient accumulation was used to achieve batch sizes 512, 1,024, and 2,048 with sub-batches of size 256.

To evaluate the performance of a Siamese network, a label of $\{abnormal,normal\}$ needs to be assigned to a patch pair. However, there is no clear cut for an uniformly sampled patch pair. Each patch pair either has no overlap or partially overlaps with a ROI. Therefore, we define the abnormal area metric $A \in[0,1]$ as follows. Let \((x_1,x_2)\) and \((y_1,y_2)\) be the coordinates of a patch and \((x_{min},x_{max})\) and \((y_{min},y_{max})\) be the bounding box coordinates. The following equation calculates $A$:
\begin{equation} \label{eq6}
\begin{split}
A = \frac{[\min(x_2,x_{max})-\max(x_1,x_{min})]\cdot[\min(y_2,y_{max})-\max(y_1,y_{min})]}{\min([(x_2-x_1)\cdot(y_2-y_1)],[(x_{max}-x_{min})\cdot(y_{max}-y_{min})])}
\end{split}
\end{equation}
This metric captures the amount of overlap between a patch and a ROI, divided by the smaller of the two areas. An AUC of a model's capability to distinguish abnormal from normal patch pairs can be calculated when $A$ is set at any cutoff $\in[0,1]$. We vary the cutoff at $100$ uniform steps in $(0,1)$ and report the average AUC. 

At the conclusion of the above mentioned grid search, we did not find any trend with respect to batch size. Therefore, Table 1 reports only the best combination of batch size and learning rate on the validation datasets as well as the corresponding test set performance. Overall, the model performed better on the patch pair datasets from the VinDr image dataset than the OPTIMAM dataset. The best results were achieved using the VinDr-256 dataset at a batch size of 2,048 and a learning rate of $1.0\times10^{-3}$ with an average AUC of 0.722 and 0.728 on the validation and test sets respectively. The model had a lower performance on the VinDr-96 dataset at a batch size of 1,024 and learning rate of $1.0\times10^{-7}$ with an average AUC of 0.673 and 0.689 on the validation and test sets. Our model performed slightly worse on the OPTIMAM-96 dataset at a batch size of 512 and a learning rate of $1.0\times10^{-5}$ with an average AUC of 0.651 and 0.569 on the validation and test sets. For the OPTIMAM-256 dataset, the best combination with a  batch size of 256 and learning rate of $1.0 \times 10^{-6}$ only achieved an average performance of 0.578 on the validation set and 0.531 on the test set. The models that achieved the best validation performance on each paired patch dataset were saved and used on downstream tasks.

\begin{table}[ht]
    \caption{Best validation set performances from the grid search for the optimal batch size and learning rate by dataset. The corresponding test set performance is also reported. Abnormal vs. normal classification of uniformly tiled mammogram patches was done with labels $\{0,1\}$ set at various abnormal area $A$ cutoff values and the AUC evaluated. The reported AUC value is the average across all cutoffs used.}
    \centering
    \begin{adjustbox}{width=\columnwidth,center}
    \begin{tabular}{ccccc}
            \toprule
            \cmidrule(r){1-5}
            Dataset & Batch Size & Learning Rate & Average Validation AUC & Average Test AUC\\
            \midrule
            VinDr-256 & 2,048 & \(1.0\times10^{-3}\) & 0.722 & 0.728\\
            
            VinDr-96 & 1,024 & \(1.0\times10^{-7}\) & 0.673 & 0.689\\
            
            OPTIMAM-96 & 512 & \(1.0\times10^{-5}\) & 0.651 & 0.569\\
            
            OPTIMAM-256 & 256 & \(1.0\times10^{-6}\) & 0.578 & 0.531\\
            \bottomrule
    \end{tabular}
    \end{adjustbox}
    \label{tab:pairedvalidation}
\end{table}

To demonstrate the training process of our models, Figure 4 shows the GMMs at the end of training as well as some example patch pairs in the VinDr-256 patch dataset. The distribution of the Euclidean distances of patch pairs: $D_1$ and $D_2$ in the histograms of Figure 4a show heavy right tails that correspond to the second component of the GMMs. This component represents the patch pairs with higher Euclidean distances, hypothetically the abnormal class. 
Figure 4b shows a couple of true abnormal patch pair examples that have high Euclidean distances $(D_1,D_2)$, high posterior probabilities ($P_1,P_2)$, and low similarity predictions $(q_1,q_2)$. For abnormal pair \#846, very different embeddings are encoded as the distances of this pair are $D_1 = 14.86$ and $D_2 = 13.81$. These extremely high distance values make the pair firmly belong to the second component of the GMMs, therefore the posterior probabilities for this pair are $P_1 = P_2 = 1.00$. The high distances imply the encoders produce different embeddings for the patch pair, hence the two networks' similarity predictions are low at $q_1 = 0.074$ and $q_2 = 0.297$. For the abnormal pair \#2,372, the networks do not make as great predictions, but they still show success. Since the distances of the embeddings for each network are relatively lower at $D_1 = 6.25$ and $D_2 = 7.05$, the posterior probabilities for the pair are also lower with $P_1 = 0.632$ and $P_2 = 0.925$. Network 1 returns a low similarity prediction of $q_1 = 0.149$ and network 2 returns a higher prediction of $q_2 = 0.394$. 
In Figure 4c, we illustrate some normal patch pairs that are incorrectly identified as abnormal. Normal pair \#0 has moderately large distances of $D_1 = 6.42$ and $D_2 = 7.12$, low similarity probabilities of $q_1 = 0.0747$ and $q_2 = 0.223$, and high posterior probabilities of $P_1 = 0.721$ and $P_2 = 0.938$. Similarly, normal pair \#1219 returns distances $D_1 = 6.29$ and $D_2 = 6.86$, high posterior probabilities $P_1 = 0.660$ and $P_2 = 0.874$. Interestingly, network 1's similarity probability is low with $q_1 = 0.383$, while network 2's similarity probability is decently high at $q_2 = 0.610$. From examining these normal patch pairs, we speculate that the networks are sensitive to visual differences in a pair. These differences may not be solely attributed to the presence of a lesion, potentially resulting in false positives.

 \begin{figure}
    \hspace*{-.1in}
    \centering
    \begin{tabular}{cc}
    \adjustbox{valign=b}{\subfloat[Histogram of the Euclidean distances of training patch pairs and GMM curve fits at the end of training. The prior probabilities of the two components are shown in the legend.\label{subfig-1:dummy}]{%
          \includegraphics[width=.47\linewidth,height=10cm]{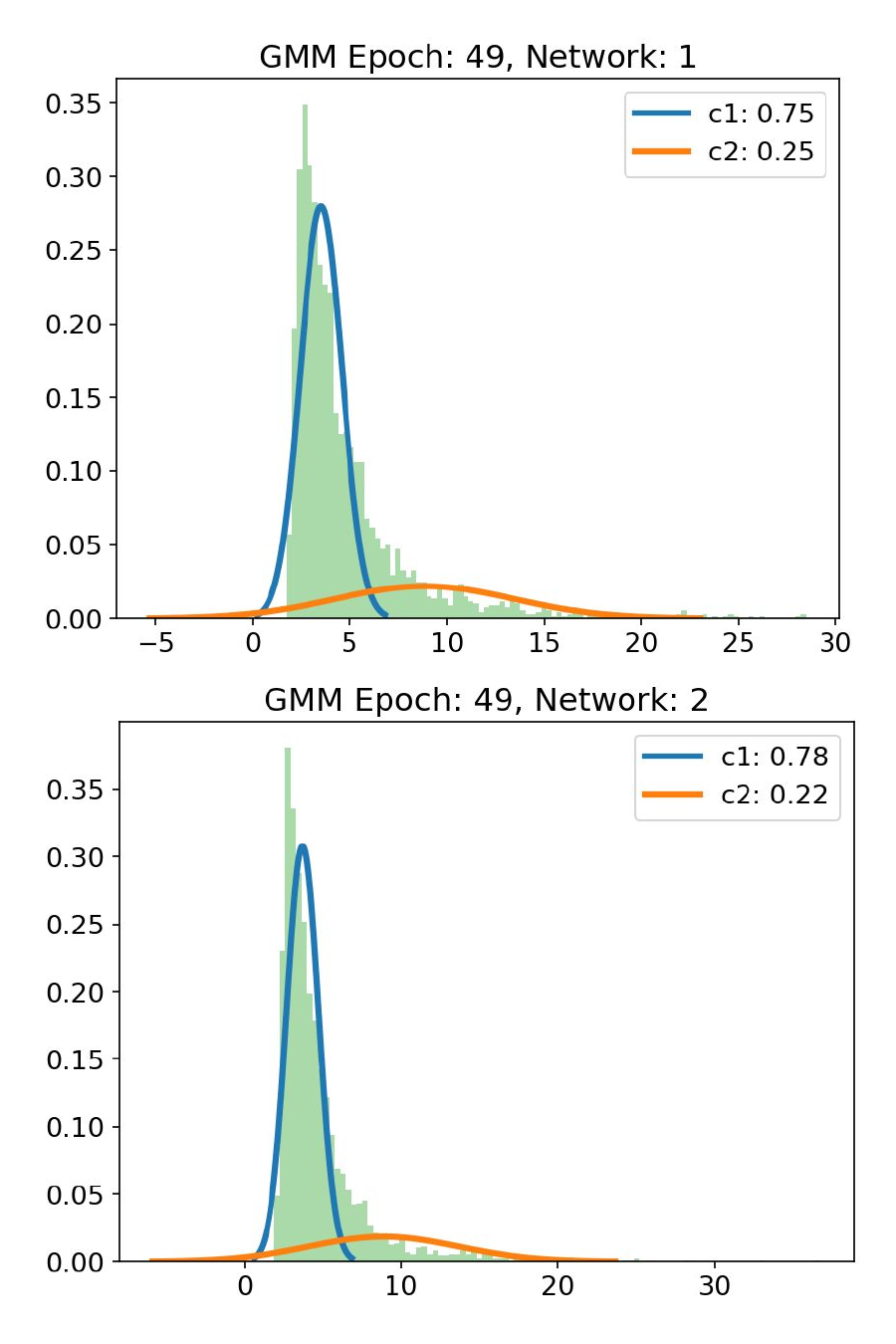}}}
    &      
    \adjustbox{valign=b}{\begin{tabular}{@{}c@{}}
    \subfloat[Abnormal patch pair samples \#846 (upper) \#2,372 (lower) are successfully predicted as abnormal by the GMM and receive low similarity predictions by the Siamese networks.\label{subfig-2:dummy}]{%
          \includegraphics[width=.5\linewidth]{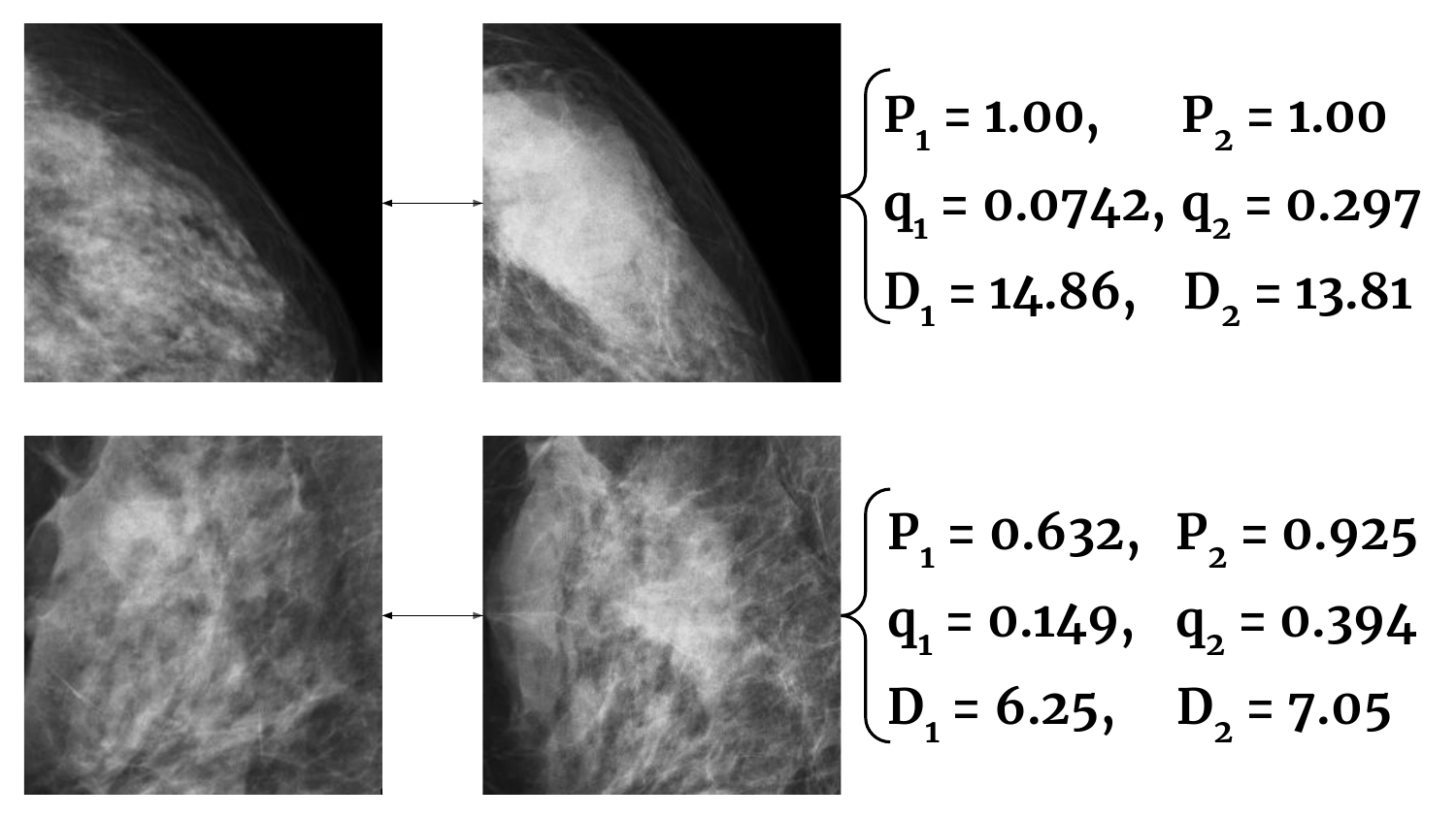}} \\
    \subfloat[Normal patch pair samples \#0 (upper) and \#1,219 (lower) are incorrectly predicted as abnormal by the GMM and assigned low similarity probabilities by the Siamese networks.\label{subfig-3:dummy}]{%
          \includegraphics[width=.5\linewidth]{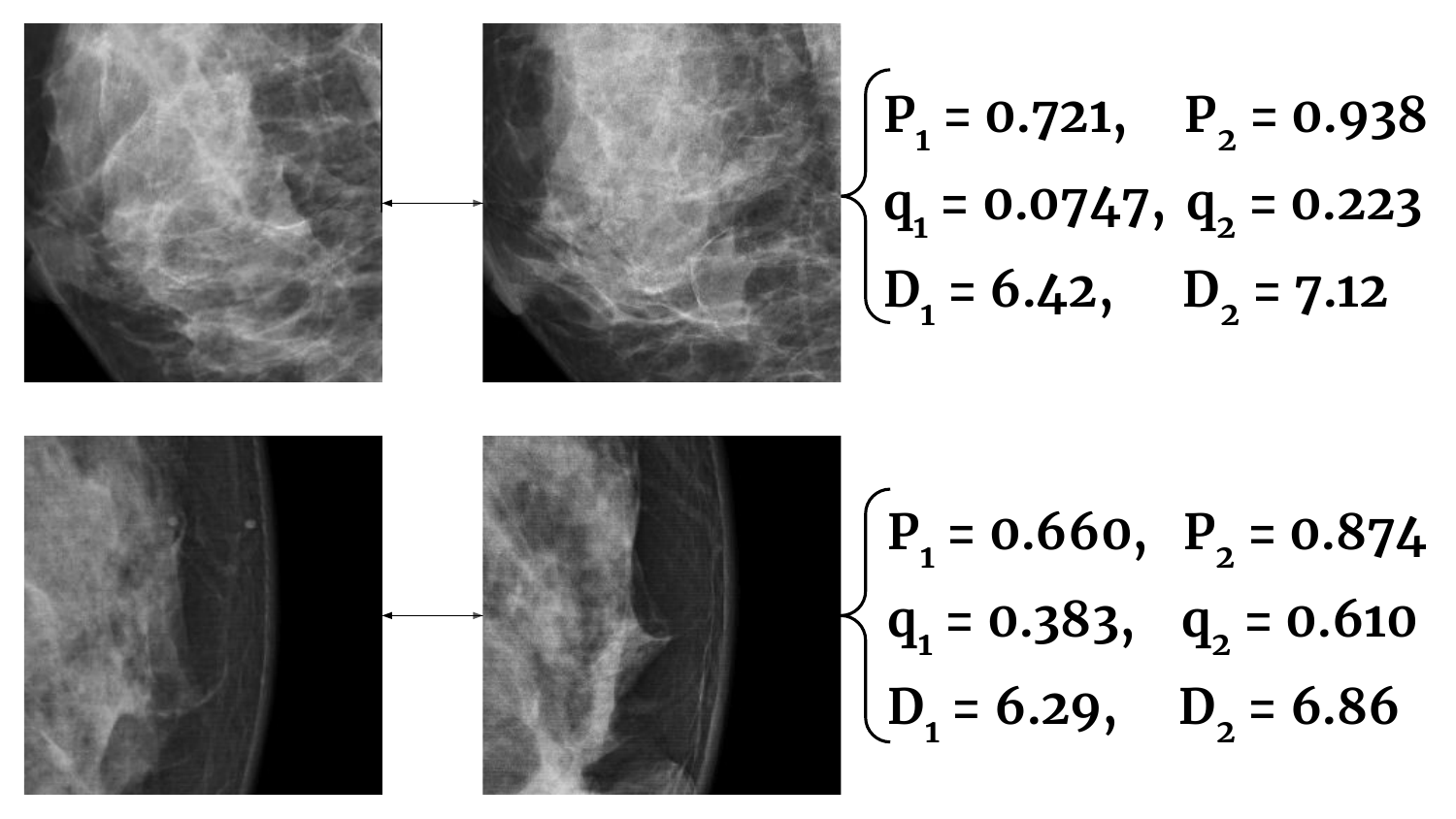}}
    \end{tabular}}
    \end{tabular}
    \caption{The GMMs at the conclusion of training on the VinDr-256 paired patch dataset. Provided are some examples of abnormal patch pairs that are correctly identified and normal patch pairs that are incorrectly identified as abnormal by the model. The posterior probability $P$, Siamese network similarity probability $q$, and patch pair Euclidean distance $D$ of each network for the patch pairs are shown.}\label{fig:dummy}
  \end{figure}
  
\subsection{Patch Pair Embedding Analysis}

To further explore the Siamese networks' embeddings of these patch pairs, we use dimension reduction methods to visualize the concatenated embeddings $E$ (size=$1024$) on 2D plots. The pairs are colored differently based on the abnormal area metric $A$, with $A = 0$ being a normal pair, $A \in (0,0.5]$ being modest overlap with ROI or $A \in (0.5,1.0]$ being high overlap with ROI. This categorization helps us understand whether the networks are able generate meaningful embeddings that can distinguish lesions from normal tissues. We use two different methods to achieve this: t-Distributed Stochastic Neighbor Embedding (t-SNE) \cite{tsne} and Uniform Manifold Approximation and Projection (UMAP) \cite{umap}. t-SNE tends to do well with preserving local structure while UMAP has the ability to preserve both local and global structure in 2D projections.

t-SNE is used to visualize the concatenated embeddings $E$ of 10,000 patch pairs from each of the paired patch datasets in the two-dimensional space. The sklearn package \cite{scikit-learn} is used for t-SNE. Figure 5 shows the t-SNE plots of the sampled patch pairs from each patch pair dataset and their corresponding label determined by $A$. In the VinDr-96 t-SNE plot in Figure 5b, there is a large clustering of samples with $A > 0.5$ in the lower half of the graph. There is also a gradual weaker association of samples with $0 < A \leq 0.5$ above. Both of these clusters mostly overlap with each other but show distinction with the normal class. Figure 5a shows more sporadic clustering of VinDr-256 patch pair samples with $A \neq 0$. Smaller clusters of samples with $A > 0.5$  and $0 < A \leq 0.5$ can be observed but there is no large cluster that represents the majority of the samples in the two classes. Though the OPTIMAM-96 patch dataset contains less abnormal patch pairs, Figure 5d demonstrates the model's ability to distinguish most of these abnormal samples within a cluster in the upper left quadrant. Since our models perform the worst on the OPTIMAM-256 dataset, it is no surprise that there is little association to be drawn in its t-SNE plot in Figure 5c.

UMAP is used to visualize the same embeddings and shows more success in Figure 6. Figure 6a shows a strong clustering of abnormal samples with $A > 0.5$ in the upper left and lower left quadrants. Among them are many samples of the $0 < A \leq 0.5$ class, but there is another small clustering of these samples in the upper right quadrant. Figure 6b also shows a strong association of samples in the $A > 0.5$ and $0 < A \leq 0.5$ classes. Unsurprisingly, the projection of the OPTIMAM-256 patch pair embeddings in Figure 6c still show little association of samples within the same class or even different classes. Though there are proportionally fewer abnormal pairs in the OPTIMAM-96 dataset, the UMAP projection of these embeddings shows a better clustering of them in the upper left quadrant. Even though a strong grouping of the abnormal patch pairs can be shown, there are still some normal pairs within these clusters. High false positive rates seem to be an issue with our model and these plots demonstrate how prevalent the false positive samples are.

Through examining the t-SNE and UMAP projections of the patch pair embeddings, our models show the ability to distinguish abnormal pairs from majority of the normal pairs in the training sets. However, cluster distinction is not very strong and there is a considerable amount of false positives. This indicates that even though there is no lesion present in these normal patches, differences in breast tissues could contribute to false classification of the normal patch pairs.

\begin{figure}
    \begin{subfigure}[b]{0.5\textwidth}
        \includegraphics[scale=.5]{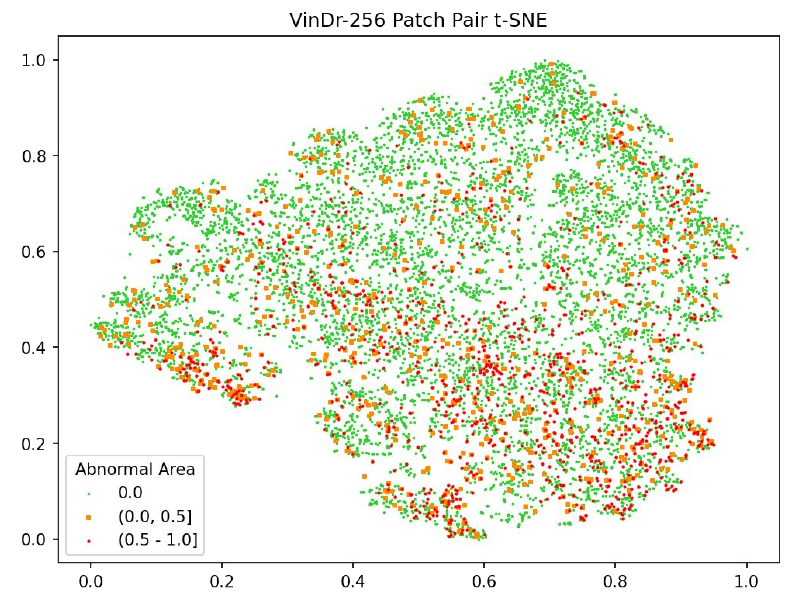}
        \caption{VinDr-256 t-SNE}
        \label{fig:v256tsne}
    \end{subfigure}
    \hfill
    \begin{subfigure}[b]{0.5\textwidth}
        \includegraphics[scale=.5]{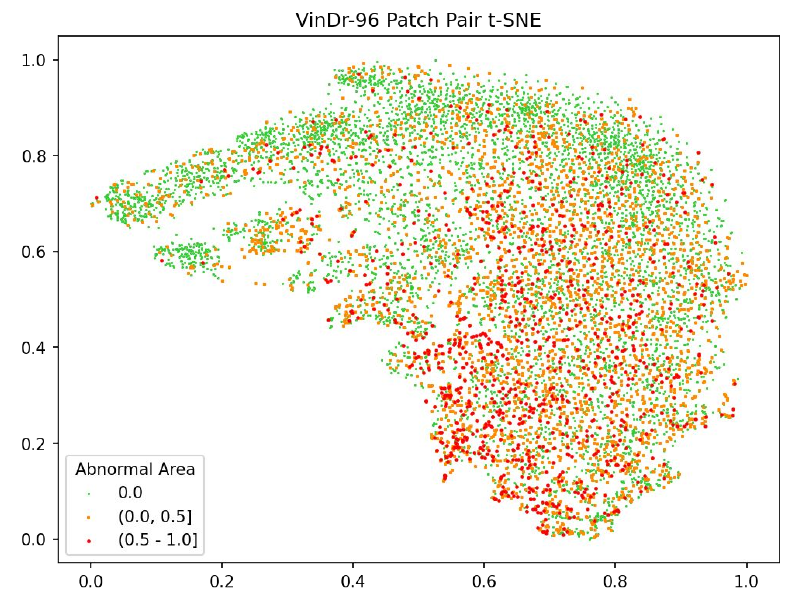}
        \caption{VinDr-96 t-SNE}
        \label{fig:v96tsne}
    \end{subfigure}
    \vfill
    \begin{subfigure}[b]{0.5\textwidth}
        \includegraphics[scale=.5]{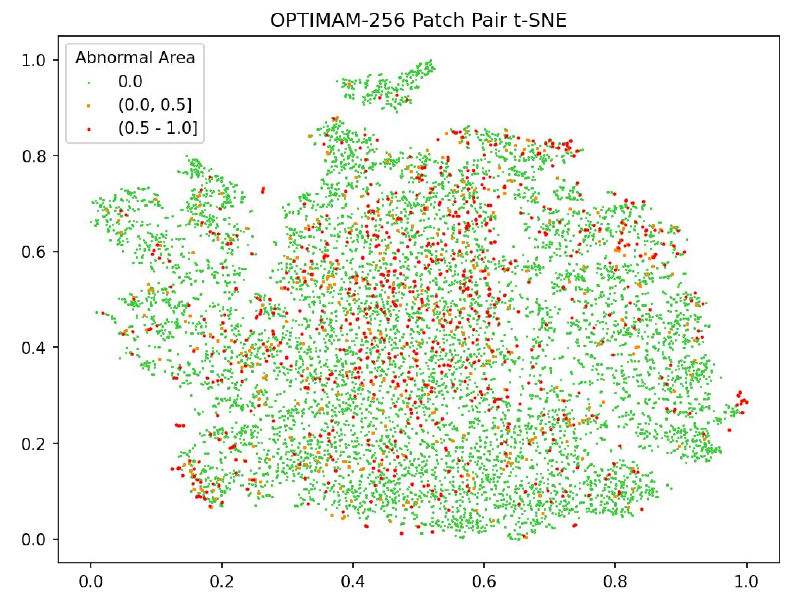}
        \caption{OPTIMAM-256 t-SNE}
        \label{fig:o256tsne}
    \end{subfigure}
    \hfill
    \begin{subfigure}[b]{0.5\textwidth}
        \includegraphics[scale=.5]{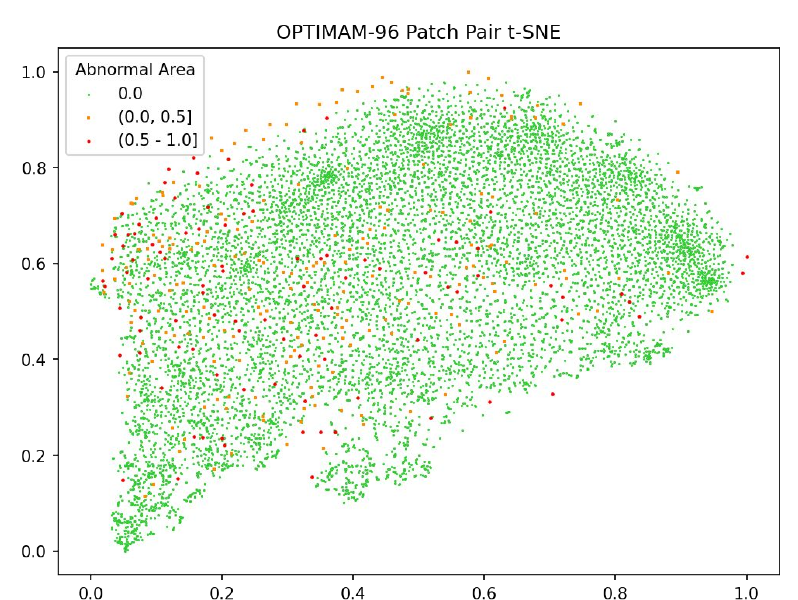}
        \caption{OPTIMAM-96 t-SNE}
        \label{fig:o96tsne}
    \end{subfigure}
    
    \label{fig:tsne}
    \caption{t-SNE plots of 10,000 samples in the VinDr-256, VinDr-96, OPTIMAM-256, and OPTIMAM-96 paired patch datasets labeled by the proportion of abnormal area $A$.}
\end{figure}

\begin{figure}
    \begin{subfigure}[b]{0.5\textwidth}
        \includegraphics[scale=.5]{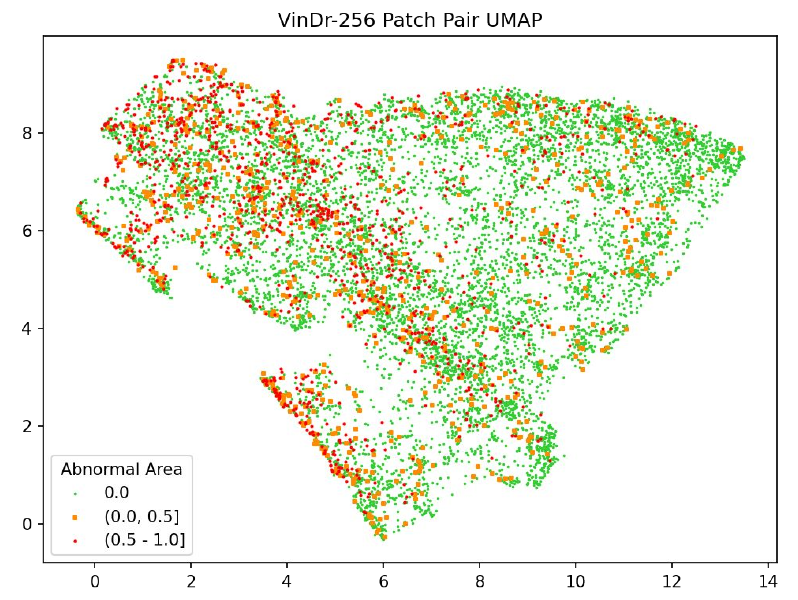}
        \caption{VinDr-256 UMAP}
        \label{fig:v256umap}
    \end{subfigure}
    \hfill
    \begin{subfigure}[b]{0.5\textwidth}
        \includegraphics[scale=.5]{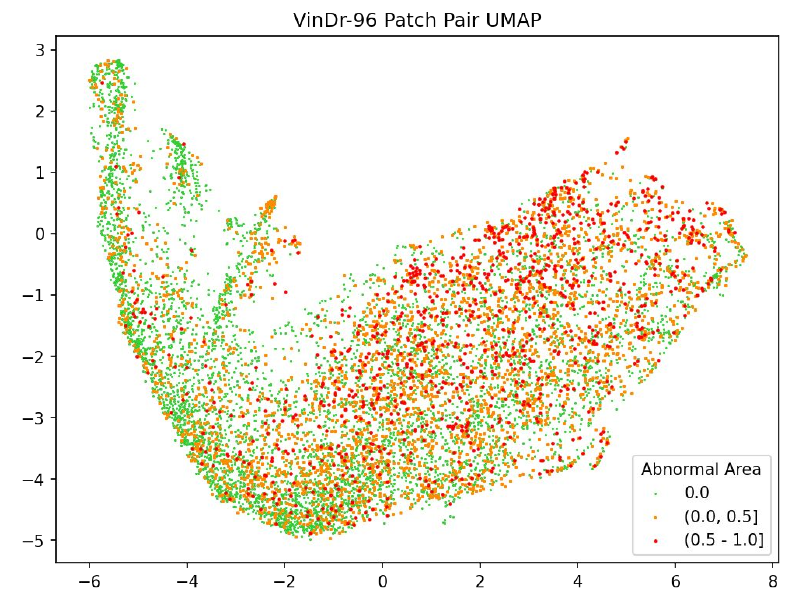}
        \caption{VinDr-96 UMAP}
        \label{fig:v96umap}
    \end{subfigure}
    \vfill
    \begin{subfigure}[b]{0.5\textwidth}
        \includegraphics[scale=.5]{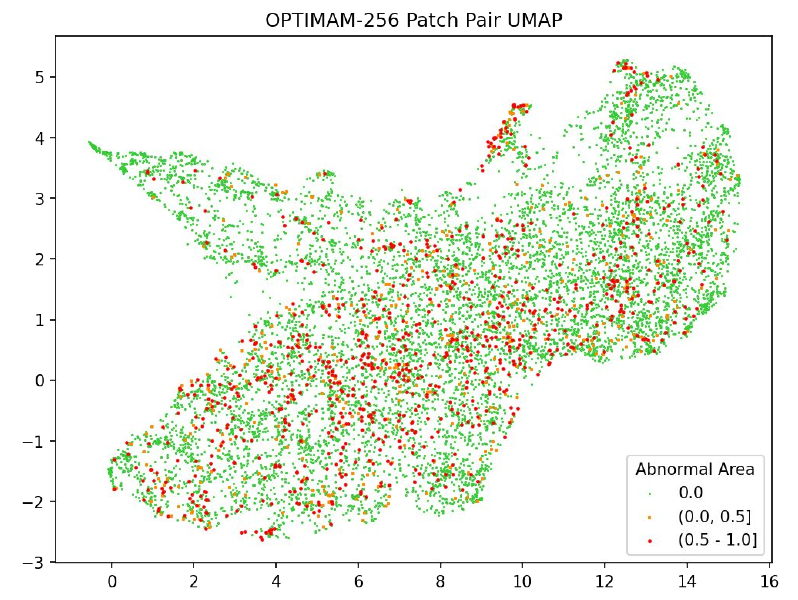}
        \caption{OPTIMAM-256 UMAP}
        \label{fig:o256umap}
    \end{subfigure}
    \hfill
    \begin{subfigure}[b]{0.5\textwidth}
        \includegraphics[scale=.5]{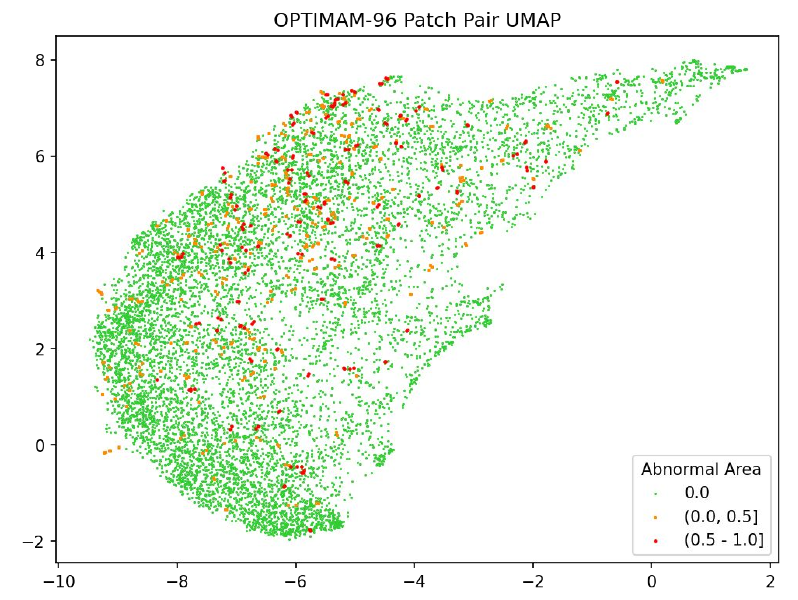}
        \caption{OPTIMAM-96 UMAP}
        \label{fig:o96umap}
    \end{subfigure}
    
    \label{fig:umap}
    \caption{UMAP plots of 10,000 samples in the VinDr-256, VinDr-96, OPTIMAM-256, and OPTIMAM-96 paired patch datasets labeled by the proportion of abnormal area $A$.}
\end{figure}

\subsection{SSL Training}

Two popular SSL methods were used as baselines for the proposed model: SimCLR and BYOL. We used the same mammogram-specific transformations as in a previous study \cite{miller}: random crop with resizing, gamma shift, and contrast. Each model was trained with a ResNet-18 encoder for 100 epochs at a batch size of 2,048. Due to computational resource constraints, gradient accumulation was used to achieve this batch size with 8 sub-batches of size 256. The default parameters of each method were used as well. The LARS \cite{lars} optimizer with a base learning rate that is linearly scaled by the batch size was used. For SimCLR, we used their learning rate scaling policy of $\frac{0.3 \times 2048}{256} = 2.4$. In BYOL, the learning rate is slightly different at $\frac{0.2 \times 2048}{256} = 1.6$. The models with the lowest validation loss on each dataset were saved during training to later be used for downstream tasks. Since SSL methods can only tell whether the two views from the same patch are different or not, we are not able to produce AUCs for the SSL training.

\subsection{Downstream Task Results}

To evaluate the pretrained models, the standard linear evaluation protocol was used, i.e. froze the image encoder's parameters and trained a linear classifier on the embeddings. For each Siamese network model, there are two separate encoders trained in parallel. We added a linear classifier on top of each encoder and used the average output as the ensemble's prediction. The linear classifiers for the Siamese, BYOL, and SimCLR pretrained encoders were trained for 100 epochs at a batch size of 32, a learning rate of  0.01, and weight decay of $10^{-5}$ with the Adam \cite{adam} optimizer. Three downstream tasks were designed to evaluate the effectiveness of the pretrained models. The AUC for both binary and multiclass classification tasks are reported. For calculating multiclass AUC, we adopt the OvR strategy (one versus rest) to evaluate the models' ability to distinguish between multiple classes. The OvR strategy involves treating each class as it's own binary classification task, where the class of interest is the positive class and all others are the negative class. This allows us to analyze the model's performance on each class as well as the overall average of the binary AUC scores across all classes.

The first task is the binary classification of abnormal versus normal patches. Table 2 reports the AUC of this classification task on all pretraining methods and image datasets. On VinDr-256 labeled dataset, the Siamese model performed the best with an AUC of 0.927, followed by the BYOL model with an AUC of 0.908, and then SimCLR with an AUC of 0.737. With the VinDr-96 labeled dataset, the Siamese pretrained model performed the best at 0.869, followed by BYOL at 0.856, and then SimCLR at 0.820. Generally, all pretraining methods have shown a higher performance on the VinDr datasets than the OPTIMAM datasets. On OPTIMAM-256 labeled dataset, the Siamese model achieved the highest AUC at 0.830, followed by BYOL at 0.782, and then SimCLR at 0.733. For the OPTIMAM-96 labeled dataset, Siamese performed best at an AUC of 0.820, followed by the BYOL model at 0.813, and once again SimCLR with the lowest AUC of 0.798. When comparing each model's performance on different patch sizes, neither patch size consistently outperforms the other. Overall, the Siamese model is either on par or better than the two SSL models.
\begin{table}
    \centering
    \caption{Test set AUC of linear evaluation of Siamese, BYOL, and SimCLR pretrained models on abnormal versus normal patch classification task.}
    \begin{tabular}{ccc}
    \toprule
    \cmidrule(r){1-3}
        Dataset & Model & AUC\\
        \midrule
         VinDr-256& Siamese & 0.927\\
         & BYOL & 0.908\\
         & SimCLR & 0.737\\
         \cmidrule(r){1-3}
         VinDr-96 & Siamese & 0.869\\
         & BYOL & 0.856\\
         & SimCLR & 0.820\\
         \cmidrule(r){1-3}
         OPTIMAM-256& Siamese & 0.830\\
         & BYOL & 0.782\\
         & SimCLR & 0.733 \\
         \cmidrule(r){1-3}
         OPTIMAM-96 & Siamese & 0.820\\
         & BYOL & 0.813\\
         & SimCLR & 0.798 \\
         \bottomrule
    \end{tabular}
    \label{tab:abvnresults}
\end{table}

The second task is BI-RADS classification of categories 1, 3, 4, and 5 on the VinDr datasets. BI-RADS 2 was excluded. Table 3 reports the average multi-class AUC along with the AUC of each class using the OvR strategy. On the VinDr-256 labeled dataset, the Siamese pretrained model performed the best with an average AUC of 0.784, followed by BYOL at 0.760, then SimCLR at 0.594. For the VinDr-96 patch set, the BYOL pretrained model performs the best with an AUC of 0.798, followed by the Siamese model at 0.765, and finally SimCLR at 0.708. Here we note a slight increase in performance when increasing the patch size in Siamese pretrained models, but this does not hold for the BYOL or SimCLR models.

\begin{table}
     \caption{VinDr test set AUC of Siamese, BYOL, and SimCLR pretrained models linear evaluation with the BI-RADS classification task. The average multiclass AUC is reported along with a breakdown of each binary AUC per class using the OvR (one versus rest) approach. The classes are BI-RADS 1 (B1), BI-RADS 3 (B3), BI-RADS 4 (B4), and BI-RADS 5 (B5).}
    \label{tab:biradsresults}
    \centering
    \begin{tabular}{ccccccc}
    \toprule
    \cmidrule(r){1-7}
        Dataset & Model & Average AUC & B1 vs. Rest & B3 vs. Rest & B4 vs. Rest & B5 vs. Rest\\
        \midrule
         VinDr-256 & Siamese & 0.784 & 0.930 & 0.737 & 0.639 & 0.830\\
         & BYOL & 0.760 & 0.929 & 0.717 & 0.616 & 0.776\\
         & SimCLR & 0.594 & 0.624 & 0.545 & 0.502 & 0.707\\
         \midrule
         VinDr-96 & Siamese & 0.765 & 0.900 & 0.697 & 0.689 & 0.773\\
         & BYOL & 0.798 & 0.921 & 0.740 & 0.668 & 0.864\\
         & SimCLR & 0.708 & 0.811 & 0.691 & 0.577 & 0.752\\
         \bottomrule
    \end{tabular}

\end{table}

The last task is 3-way classification of background, benign and malignant on the OPTIMAM dataset. Table 4 reports the average multi-class AUC along with the AUC of each class using the OvR strategy. On OPTIMAM-256 patch dataset, the Siamese model performed best at 0.744, followed by BYOL at 0.679, and then SimCLR at 0.643. With the OPTIMAM-96 labeled dataset, the Siamese model achieves the best performance at 0.732, slightly following behind is BYOL at 0.722, and once again in last is SimCLR at 0.719. Models pretrained on smaller patch size perform either on par or better than those pretrained with larger patches.

\begin{table}
    \caption{OPTIMAM test set AUC of Siamese, BYOL, SimCLR pretrained models linear evaluation with the outcome classification task. The average multiclass AUC is reported along with a breakdown of each binary AUC per class using the OvR (one versus rest) approach. The classes are background (N), benign (B), and malignant (M).}
    \label{tab:outcomeresults}
    \centering
    \begin{tabular}{ccccccc}
    \toprule
    \cmidrule(r){1-6}
        Dataset & Model & Average AUC & N vs. Rest & B vs. Rest & M vs. Rest\\
        \midrule
         OPTIMAM-256 & Siamese & 0.744 & 0.820 & 0.614 & 0.797\\
         & BYOL & 0.679 & 0.754 & 0.553 & 0.731\\
         & SimCLR & 0.643 & 0.693 & 0.550 & 0.686\\
         \midrule
         OPTIMAM-96 & Siamese & 0.732 & 0.824 & 0.566 & 0.806\\
         & BYOL & 0.722 & 0.801 & 0.587 & 0.779\\
         & SimCLR & 0.719 & 0.787 & 0.592 & 0.778\\
         \bottomrule
    \end{tabular}

\end{table}

\subsection{Alternative Designs}

An alternate loss function that can be used for Siamese Networks is the Triplet loss which was first introduced in FaceNet \cite{facenet}. The goal of the Triplet loss function is to minimize the distance between an anchor sample and positive samples, or similar instances, while maximizing the distances between the anchor sample and negative samples, or different instances. The positive and negative pairs must also maintain a certain distance apart denoted by margin, $m$. This setup requires prior knowledge about the samples' labels in order to correctly designate these positive and negative pairs. In our unsupervised setting, the labels are unknown for the patch pairs in the training set. To accommodate this, we used soft label $P$ and $1-P$ as weights on $D$ to split it into the distances for the negative and positive pairs. That is, $D_{+}=(1-P) \cdot D$ represents the distance of a positive pair and $D_{-}=P \cdot D$ represents the distance of a negative pair. After making the appropriate changes, the following loss function was used:

\begin{equation} \label{eq5}
\begin{split}
L = [(1-P) \cdot D - P \cdot D+m]_{+}
\end{split}
\end{equation}

\noindent We experimented with different margin values $m$,
but overall this method proved to be unstable in training and would yield different results per run.

\par We also explored a slightly different approach to derive the soft label $P$. Recall from Figure 1 that in the Siamese network, patch pair $(p_1,p_2)$ is encoded to embeddings $(e_1,e_2)$, which are then concatenated to a single vector $E$, and finally passed through a linear layer to a single output node. Let the output of this node before applying the softmax activation function be denoted as $z$. Instead of fitting the GMM function $h$ to the Euclidean distance set $C$, we experimented with fitting $h$ to the linear output set $Z = \{z_i\}, i = 1..N$. A higher $z$ means the pair is more likely to be normal. The GMM function $h$ now provides the posterior probability that a pair with linear output $z$ belongs to the normal class, such that the soft label $P = 1 - h(z)$. Similarity probability $q$ is still obtained by applying the softmax activation function on $z$. Equations 2-4 remain unchanged. Our efforts include training a single Siamese network and the double Siamese network models. However, the validation performances were extremely volatile throughout training. Therefore, we deemed this method to be too unstable to move forward with.

Departing from using Siamese networks, we originally focused more on a SSL-style approach. Given a chronological mammogram patch pair $(p_1,p_2)$ we used strong data augmentation on each image to produce views $(v_{11},v_{12})$ from $p_1$ and $(v_{21},v_{22})$ from $p_2$. Then the loss values of each combination of views are obtained using the SSL method. In total, there are the loss values of views originating from the same patch, $l(v_{11},v_{12})$ and $l(v_{21},v_{22})$, and the losses of views from different patches, $l(v_{11},v_{21})$, $l(v_{11},v_{22})$, $l(v_{12},v_{21})$, and $l(v_{12},v_{22})$. In our model's loss function, we scale the average SSL losses of views from the same patch and the average SSL losses of views from different patches with soft label $P$ and $1-P$, respectively, which is still obtained using a GMM function $h$. $h$ is fit to the set of SSL losses of untransformed patch pairs in the training set. Here $P = h(l(p_1,p_2))$, represents the posterior probability that patch pair $(p_1,p_2)$ with a SSL loss value of  $l$, is an abnormal pair. Equation 7 is the model's final loss function:

\begin{equation}
    \begin{multlined}
    \hspace*{-0.2in} 
    L = P\cdot\frac{l(v_{11},v_{12})+l(v_{21},v_{22})}{2} \\  \hspace*{0.8in} +(1-P)\cdot\frac{l(v_{11},v_{21})+l(v_{11},v_{22})+l(v_{12},v_{21})+l(v_{12},v_{22})}{4}
    \end{multlined}
\end{equation}
\raggedbottom
The first part of the loss is weighted by $P$ to represent the portion of the pair that is abnormal (therefore, we look at only the two views from the same patch); the second part of the loss is weighted by $1-P$ to represent the portion of the pair that is normal (therefore, we look at cross-views from the two patches). We tried using several SSL methods including SimCLR and BYOL, but training was not successful. Upon examination of the GMM plots, the distribution of losses of the untransformed patch pairs did not follow either a bimodal distribution or a normal distribution. The GMM was unable to capture this behavior and therefore soft label $P$ prediction was inaccurate.

\section{Discussion and Conclusion}

In this study, an algorithm that utilizes a Siamese neural network with soft labels is developed to assess the similarity of bilateral mammogram patch pairs without supervision. An encoder is trained with the aim to generate the same embeddings for similar pairs and different embeddings for abnormal pairs. A soft label is introduced for training these networks to deal with the lack of annotations. This is derived by fitting a Gaussian mixture model on the Euclidean distances of the patch pair embeddings on a training set. We found that simultaneously training two Siamese networks where the outputs were cross used in each other's loss functions showed the most success. These pretrained encoders can then be transferred for downstream tasks such as abnormal versus normal classification, BI-RADS classification, and outcome classification.

SimCLR and BYOL are two SSL methods that were used to compare with our proposed model. On all downstream tasks, the Siamese networks outperformed or performed on par with the two SSL methods. The Siamese network model shows great success in the binary abnormal versus normal patch classification task compared with the SSL pretraining methods. This performance is attributed to the design of the Siamese network pretraining to distinguish bilateral patch pairs. This is also supported by the embedding analysis using t-SNE and UMAP, where the Siamese networks show the ability to detect these abnormal patch pairs among an abundance of normal pairs. However these clusters are not perfect as it shows many normal pairs remain in these abnormal clusters, supporting the prevalence of false positives identified by the model. 

When further evaluating model performance on more difficult classification tasks by splitting the abnormal class into more categories, we gained insight on the type of lesions the model can confidently identify. In the BI-RADS classification task, of the abnormal classes BI-RADS 3-5, the Siamese networks are best at distinguishing lesions originating from BI-RADS 5 images. Also, in the OPTIMAM patient outcome classification task, the Siamese networks perform well in distinguishing malignant patches from benign and background patches. This suggests that the Siamese network was able to learn features of malignant lesions without being explicitly given this information. Ideally, we would also want the Siamese network to be able to detect less obvious lesions, as early detection is a critical part of breast cancer survival \cite{earlydetection}. 

It is important to note that data leakage might be a potential issue in these experiments. For datasets used in pretraining the split was done at patient level, while for datasets used in downstream tasks the split was done at patch level. Although the patches from the two sets don't overlap with each other due to different sampling methods, studies from patients used in the training set of pretraining may appear in the test sets of downstream patch datasets.

Further research and potential applications of this work need to be explored. This study focuses in the scope of patch pretraining to patch classification. More complicated downstream tasks such as entire mammogram image classification can prove to be more applicable for a clinical use. Additionally, more methods for deriving the soft label should be explored as well, considering it is the backbone of this unsupervised method. An interesting observation is that the histogram distribution of embedding distances $D$, always show a heavy right tail. While we were not surprised by this behavior, the distribution of these higher distances do not exactly follow a perfect normal distribution. This was a major assumption that the distribution of distances would behave this way so perhaps a different mixture modeling method could be explored. A beta distribution is more flexible in modeling skewed distributions, so utilizing a Beta mixture model could be an alternative to better capture this behavior and therefore make more accurate soft label predictions during training. \cite{ICML2019_UnsupervisedLabelNoise}. Since this algorithm relies on symmetrical image inputs, applications on different medical imaging datasets that have this feature should be also considered.

\par In short, the Siamese network model shows potential in being a powerful algorithm that can effectively be used for pretraining. This study shows that by leveraging the symmetry of the human body, an encoder can be trained to identify the presence of abnormalities in mammogram patches. By pretraining on an abundance of patch pairs in an unsupervised manner, a flexible and reliable encoder can be used on a variety of downstream tasks.

\section{Acknowledgements}
The images and data used in this publication are derived from the OPTIMAM imaging database (https://medphys.royalsurrey.nhs.uk/omidb/) \cite{OPTIMAM}, we would like to acknowledge the OPTIMAM project team and staff at the Royal Surrey County Hospital who developed the OPTIMAM database, and Cancer Research UK who funded the creation and maintenance of the database.

This work was supported in part through the computational and data resources and staff expertise provided by Scientific Computing and Data at the Icahn School of Medicine at Mount Sinai and supported by the Clinical and Translational Science Awards (CTSA) grant UL1TR004419 from the National Center for Advancing Translational Sciences.
\pagebreak
\AtNextBibliography{\small}
\printbibliography

@article{simclr,
  title={A Simple Framework for Contrastive Learning of Visual Representations},
  author={Chen, Ting and Kornblith, Simon and Norouzi, Mohammad and Hinton, Geoffrey},
  journal={arXiv preprint arXiv:2002.05709},
  year={2020}
}

@article{byol,
    title={Bootstrap Your Own Latent: A New Approach to Self-Supervised Learning},
    author={Jean-Bastien Grill, and Florian Strub, and Florent Altché, and Corentin Tallec, and Pierre H. Richemond, and Elena Buchatskaya, and Carl Doersch, and Bernardo Avila Pires, and Zhaohan Daniel Guo, and Mohammad Gheshlaghi Azar, and Bilal Piot, and Koray Kavukcuoglu, and Rémi Munos, and Michal Valko},
    journal={arXiv preprint arXiv:2006.07733},
    year={2020}
}

@article{miller,
    title={Self-Supervised Deep Learning to Enhance Breast Cancer Detection on Screening Mammography},
    author={John D. Miller, and Vignesh A. Arasu, and Albert X. Pu, and Laurie R. Margolies, and Weiva Sieh, and Li Shen},
    journal={arXiv preprint arXiv:2203.08812},
    year={2022}
}

@article{facenet,
    title={FaceNet: A Unified Embedding for Face Recognition and Clustering},
    author={Florian Schroff and Dmitry Kalenichenko and James Philbin},
    journal={arXiv preprint arXiv:1503.03832},
    year={2015}
}

@article{featurefusion,
author = {Bai, Jun and Jin, Annie and Wang, Tianyu and Yang, Clifford and Nabavi, Sheida},
title = {Feature fusion Siamese network for breast cancer detection comparing current and prior mammograms},
journal = {Medical Physics},
volume = {49},
number = {6},
pages = {3654-3669},
doi = {https://doi.org/10.1002/mp.15598},
url = {https://aapm.onlinelibrary.wiley.com/doi/abs/10.1002/mp.15598},
eprint = {https://aapm.onlinelibrary.wiley.com/doi/pdf/10.1002/mp.15598},
year = {2022}
}

@inproceedings{dividemix,
title={DivideMix: Learning with Noisy Labels as Semi-supervised Learning},
author={Junnan Li and Richard Socher and Steven C.H. Hoi},
booktitle={International Conference on Learning Representations},
year={2020},
url={https://openreview.net/forum?id=HJgExaVtwr}
}

@article{ICML2019_UnsupervisedLabelNoise,
  title={Unsupervised label noise modeling and loss correction},
  author={Eric Arazo Sanchez and Diego Ortego and Paul Albert and Noel E. O’Connor and Kevin McGuinness},
  journal={ArXiv},
  year={2019},
  volume={abs/1904.11238},
  url={https://api.semanticscholar.org/CorpusID:131777002}
}

@article{scikit-learn,
 title={Scikit-learn: Machine Learning in {P}ython},
 author={Pedregosa, F. and Varoquaux, G. and Gramfort, A. and Michel, V.
         and Thirion, B. and Grisel, O. and Blondel, M. and Prettenhofer, P.
         and Weiss, R. and Dubourg, V. and Vanderplas, J. and Passos, A. and
         Cournapeau, D. and Brucher, M. and Perrot, M. and Duchesnay, E.},
 journal={Journal of Machine Learning Research},
 volume={12},
 pages={2825--2830},
 year={2011}
}

@article{Nguyen2022.03.07.22272009,
  author={Nguyen, Hieu T. and Nguyen, Ha Q. and Pham, Hieu H. and Lam, Khanh and Le, Linh T. and Dao, Minh and Vu, Van},
  title={VinDr-Mammo: A large-scale benchmark dataset for computer-aided diagnosis in full-field digital mammography},
  year={2022},
  doi={10.1101/2022.03.07.22272009},
  URL={https://www.medrxiv.org/content/early/2022/03/10/2022.03.07.22272009},
  journal={medRxiv}
}

@article{OPTIMAM,
author = {M.D., Halling-Brown and  L.M., Warren and D., Ward and E., Lewis and A., Mackenzie and M.G., Wallis and L.S., Wilkinson and R.M., Given-Wilson and R., McAvinchey and K.C., Young},
title = {OPTIMAM Mammography Image Database: A Large-Scale Resource of Mammography Images and Clinical Data},
journal = {Radiology: Artificial Intelligence},
volume = {3},
pages = {e200103},
year = {2021},
doi = {10.1148/ryai.2020200103},
    note ={PMID: 33937853},

URL = { 
    
        https://doi.org/10.1148/ryai.2020200103
    
    

},
eprint = { 
    
        https://doi.org/10.1148/ryai.2020200103
    
    

}
}

@article{cad,
author = {Chan, Heang-Ping and Hadjiiski, Lubomir M. and Samala, Ravi K.},
title = {Computer-aided diagnosis in the era of deep learning},
journal = {Medical Physics},
volume = {47},
number = {5},
pages = {e218-e227},
doi = {https://doi.org/10.1002/mp.13764},
url = {https://aapm.onlinelibrary.wiley.com/doi/abs/10.1002/mp.13764},
eprint = {https://aapm.onlinelibrary.wiley.com/doi/pdf/10.1002/mp.13764},
year = {2020}
}

@article{preparemedimg,
author = {Willemink, Martin                             J. and Koszek, Wojciech                             A. and Hardell, Cailin and Wu, Jie and Fleischmann, Dominik and Harvey, Hugh and Folio, Les                         R. and Summers, Ronald                             M. and Rubin, Daniel                             L. and Lungren, Matthew                             P.},
title = {Preparing Medical Imaging Data for Machine Learning},
journal = {Radiology},
volume = {295},
number = {1},
pages = {4-15},
year = {2020},
doi = {10.1148/radiol.2020192224},
    note ={PMID: 32068507},

URL = { 
    
        https://doi.org/10.1148/radiol.2020192224
    
    

},
eprint = { 
    
        https://doi.org/10.1148/radiol.2020192224
    
    

}
}

@INPROCEEDINGS{origsiam,
  author={Chopra, S. and Hadsell, R. and LeCun, Y.},
  booktitle={2005 IEEE Computer Society Conference on Computer Vision and Pattern Recognition (CVPR'05)}, 
  title={Learning a similarity metric discriminatively, with application to face verification}, 
  year={2005},
  volume={1},
  number={},
  pages={539-546 vol. 1},
  doi={10.1109/CVPR.2005.202}
}

@Article{sslreview,
AUTHOR = {Albelwi, Saleh},
TITLE = {Survey on Self-Supervised Learning: Auxiliary Pretext Tasks and Contrastive Learning Methods in Imaging},
JOURNAL = {Entropy},
VOLUME = {24},
YEAR = {2022},
NUMBER = {4},
ARTICLE-NUMBER = {551},
URL = {https://www.mdpi.com/1099-4300/24/4/551},
PubMedID = {35455214},
ISSN = {1099-4300},
DOI = {10.3390/e24040551}
}

@article{simpleitk,
  author={Ziv Yaniv, and Bradley C. Lowekamp, and Hans J. Johnson, and Richard Beare},
  title={SimpleITK Image-Analysis Notebooks: a Collaborative Environment for Education and Reproducible Research},
  month={11},
  day={27},
  year={2017},
  doi={https://doi.org/10.1007/s10278-017-0037-8},
  journal={Journal of Digital Imaging},
  volume={31},
  pages={290--303}
}

@article{han2020survey,
  title={A Survey of Label-noise Representation Learning: Past, Present and Future},
  author={Bo Han and Quanming Yao and Tongliang Liu and Gang Niu and Ivor Wai-Hung Tsang and James Tin-Yau Kwok and Masashi Sugiyama},
  journal={ArXiv},
  year={2020},
  volume={abs/2011.04406},
  url={https://api.semanticscholar.org/CorpusID:226282258}
}

@article{adam,
  title={Adam: A Method for Stochastic Optimization},
  author={Diederik P. Kingma and Jimmy Ba},
  journal={CoRR},
  year={2014},
  volume={abs/1412.6980},
  url={https://api.semanticscholar.org/CorpusID:6628106}
}

@misc{
lars,
title={Large Batch Training of Convolutional Networks with Layer-wise Adaptive Rate Scaling},
author={Boris Ginsburg and Igor Gitman and Yang You},
year={2018},
url={https://openreview.net/forum?id=rJ4uaX2aW},
}

@ARTICLE{ssl_general,
  author={Ericsson, Linus and Gouk, Henry and Loy, Chen Change and Hospedales, Timothy M.},
  journal={IEEE Signal Processing Magazine}, 
  title={Self-Supervised Representation Learning: Introduction, advances, and challenges}, 
  year={2022},
  volume={39},
  number={3},
  pages={42-62},
  doi={10.1109/MSP.2021.3134634}}

@article{ssl_med, title={Self-supervised learning methods and applications in Medical Imaging Analysis: A survey}, volume={8}, DOI={10.7717/peerj-cs.1045}, journal={PeerJ Computer Science}, author={Shurrab, Saeed and Duwairi, Rehab}, year={2022}}

@article{doi:10.1148/ryai.220056,
author = {Abdalla, Mohamed and Fine, Benjamin},
title = {Hurdles to Artificial Intelligence Deployment: Noise in Schemas and                     “Gold” Labels},
journal = {Radiology: Artificial Intelligence},
volume = {5},
number = {2},
pages = {e220056},
year = {2023},
doi = {10.1148/ryai.220056},

URL = { 
    
        https://doi.org/10.1148/ryai.220056
    
    

},
eprint = { 
    
        https://doi.org/10.1148/ryai.220056
    
    

}
}

@INPROCEEDINGS{5206848,
  author={Deng, Jia and Dong, Wei and Socher, Richard and Li, Li-Jia and Kai Li and Li Fei-Fei},
  booktitle={2009 IEEE Conference on Computer Vision and Pattern Recognition}, 
  title={ImageNet: A large-scale hierarchical image database}, 
  year={2009},
  volume={},
  number={},
  pages={248-255},
  doi={10.1109/CVPR.2009.5206848}}

@article{ssl_review, title={Self-supervised learning: A succinct review}, volume={30}, DOI={10.1007/s11831-023-09884-2}, number={4}, journal={Archives of Computational Methods in Engineering}, author={Rani, Veenu and Nabi, Syed Tufael and Kumar, Munish and Mittal, Ajay and Kumar, Krishan}, year={2023}, pages={2761–2775}}

@article{tsne,
author = {van der Maaten, Laurens and Hinton, Geoffrey},
year = {2008},
month = {11},
pages = {2579-2605},
title = {Viualizing data using t-SNE},
volume = {9},
journal = {Journal of Machine Learning Research}
}

@article{dl, title={Deep learning: A comprehensive overview on techniques, taxonomy, applications and Research Directions}, volume={2}, DOI={10.1007/s42979-021-00815-1}, number={6}, journal={SN Computer Science}, author={Sarker, Iqbal H.}, year={2021}}

@INPROCEEDINGS{resnet18,
  author={He, Kaiming and Zhang, Xiangyu and Ren, Shaoqing and Sun, Jian},
  booktitle={2016 IEEE Conference on Computer Vision and Pattern Recognition (CVPR)}, 
  title={Deep Residual Learning for Image Recognition}, 
  year={2016},
  volume={},
  number={},
  pages={770-778},
  doi={10.1109/CVPR.2016.90}}

@article{umap, doi = {10.21105/joss.00861}, url = {https://doi.org/10.21105/joss.00861}, year = {2018}, publisher = {The Open Journal}, volume = {3}, number = {29}, pages = {861}, author = {Leland McInnes and John Healy and Nathaniel Saul and Lukas Großberger}, title = {UMAP: Uniform Manifold Approximation and Projection}, journal = {Journal of Open Source Software} }

@article{crowdsource,
author = {Yasmin, Romena and Hassan, Md Mahmudulla and Grassel, Joshua T. and Bhogaraju, Harika and Escobedo, Adolfo R. and Fuentes, Olac},
year = {2022},
title = {Improving Crowdsourcing-Based Image Classification Through Expanded Input Elicitation and Machine Learning},
volume = {5},
journal = {Frontiers in Artificial Intelligence},
url={https://www.frontiersin.org/articles/10.3389/frai.2022.848056}, 
doi={10.3389/frai.2022.848056},
issn={2624-8212}
}

@article{earlydetection,
author = {Ginsberg, Ophira and Yip, Cheng-Har and Brooks, Ari and Cabbanes, Anna},
year = {2020},
month = {5},
day = {15},
pages = {2379-2393},
title = {Breast cancer early detection: a phased approach to implementation},
volume = {126},
number = {10},
journal = {Cancer}
}

\end{document}